\pdfoutput=1

\documentclass[11pt]{article}

\usepackage[preprint]{coling}

\usepackage{times}
\usepackage{latexsym}

\usepackage[T1]{fontenc}

\usepackage[utf8]{inputenc}

\usepackage{microtype}

\usepackage{inconsolata}

\usepackage{graphicx}
\usepackage{svg}
\usepackage{cite}
\usepackage{algorithm}
\usepackage{algorithmic}
\usepackage{amsmath}
\usepackage{tabularx}
\usepackage{multicol}
\usepackage{multirow}
\usepackage{array}
\usepackage{booktabs}
\usepackage{bbding}
\usepackage[marginal]{footmisc}
\usepackage[most]{tcolorbox}
\usepackage{colortbl} 
\usepackage{fancyhdr} 
\usepackage{geometry} 
\usepackage{float}
\usepackage{caption}
\usepackage{CJKutf8}
\usepackage{xcolor}
\usepackage{amssymb}
\usepackage{pifont}
\usepackage{ulem}

\definecolor{darkgreen}{RGB}{0,100,0} 
\definecolor{darkred}{RGB}{160,0,0} 

\floatstyle{plain} 
\restylefloat{tcolorboxfloat}

\newfloat{tcolorboxfloat}{tbp}{lop}
\floatname{tcolorboxfloat}{Table}

\title{medIKAL: Integrating Knowledge Graphs as Assistants of LLMs for Enhanced Clinical Diagnosis on EMRs}

\author{
 \textbf{Mingyi Jia\textsuperscript{1}},
 \textbf{Junwen Duan\textsuperscript{1}}\thanks{Corresponding Author. Email: \href{mailto:jwduan@csu.edu.cn}{jwduan@csu.edu.cn}.},
 \textbf{Yan Song\textsuperscript{2}},
 \textbf{Jianxin Wang\textsuperscript{1}},
\\
 \textsuperscript{1}Hunan Provincial Key Lab on Bioinformatics, \\ School of Computer Science and
Engineering, Central South University \\
 \textsuperscript{2}University of Science and Technology of China,
\\
\texttt{\{jiamingyi, jwduan\}@csu.edu.cn, clksong@gmial.com, jxwang@mail.csu.edu.cn}
}

\begin{document}
\maketitle
\begin{abstract}
Electronic Medical Records (EMRs), while integral to modern healthcare, present challenges for clinical reasoning and diagnosis due to their complexity and information redundancy. To address this, we proposed medIKAL (\textbf{I}ntegrating \textbf{K}nowledge Graphs as \textbf{A}ssistants of \textbf{L}LMs), a framework that combines Large Language Models (LLMs) with knowledge graphs (KGs) to enhance diagnostic capabilities. medIKAL assigns weighted importance to entities in medical records based on their type, enabling precise localization of candidate diseases within KGs. It innovatively employs a residual network-like approach, allowing initial diagnosis by LLMs to be merged into KG search results. Through a path-based reranking algorithm and a fill-in-the-blank style prompt template, it further refined the diagnostic process. We validated medIKAL's effectiveness through extensive experiments on a newly introduced open-sourced Chinese EMR dataset, demonstrating its potential to improve clinical diagnosis in real-world settings. The code and dataset are publicly available at \url{https://github.com/CSU-NLP-Group/mediKAL}.
\end{abstract}

\section{Introduction}
Electronic Medical Records (EMRs) are the digitized record of a patient's medical and health information and play an important role in the modern healthcare system. However, due to their complexity and information redundancy, clinical diagnosis based on EMRs extremely requires specialized medical knowledge and clinical experience. This demand has led to the development of automated methods to assist and support clinical diagnosis and decision-making.


\begin{figure}[t]
\centering
\includegraphics[scale=0.35]{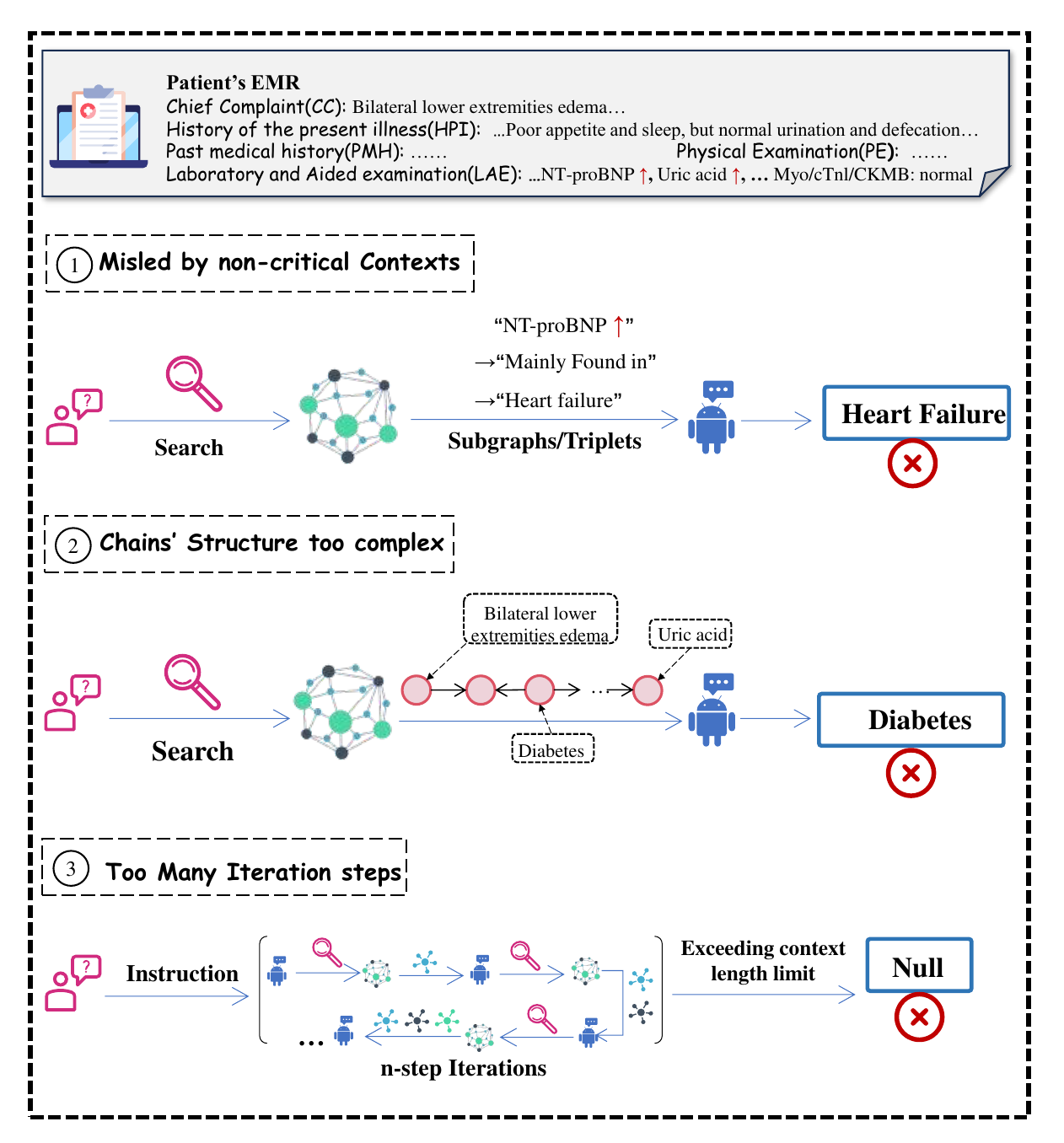}
\caption{Limitations of existing methods using KG-augmented LLMs for application to EMR diagnostic tasks. \textcircled{1} use subgraphs/triplets to augment context.\textcircled{2} use reasoning chains to augment context. \textcircled{3} use the iteration-based approach to involve LLMs in KG searching and reasoning.}
\label{limitations}
\end{figure}
Recently, large language models (LLMs) have demonstrated great potential in various medical domains~\citep{lee2023benefits,lee2023rise,ayers2023comparing,nayak2023comparison}. But directly applying LLMs to the medical field still has raised concerns about the generation of erroneous knowledge and hallucinations because of their lack of specific medical knowledge~\citep{bernstein2023comparison}. Training LLMs in the medical domain requires a lot of high-quality data, and the best-performing LLMs available are often closed-source, making further training difficult~\citep{openai2023gpt4}. Furthermore, considering that knowledge in the medical field is constantly being updated and iterated, for already trained LLMs, updating their parameters can only be done through retraining, which is extremely time-consuming and expensive~\citep{baek2023knowledge}. 

As a classic form of large-scale structured knowledge base, knowledge graphs (KGs) can provide explicit knowledge representation and interpretable reasoning paths and can be continually modified for correction or update. Therefore, KGs become an ideal complement to LLMs~\citep{pan2024unifyingLLMandKG_roadmap, yanquan2024multi}. However, existing works on "LLM$\oplus$KG" cannot be directly applied to EMR diagnosis tasks, mainly due to the following reasons: (1) Existing approaches rely on entity recognition in the input text to locate corresponding information in KGs, but they do not differentiate the contributions of different types of entities during searching on KGs. (2) They typically treat triplets or subgraphs obtained from KGs as direct context inputs or simply convert them into natural language,  which can easily lead to the problem of exceeding the input length limit and hard to understand for LLMs when encountering complex structures and informative contexts. (3) It was found that when adopting RAG paradigm, LLMs tend to overly rely on the provided context and fail to fully utilize their internal knowledge, making it easy to be misled by incorrect knowledge~\citep{rag-verify}.
\newline \indent In this paper, we propose an effective framework called medIKAL (\textbf{I}ntegrating \textbf{K}nowledge Graphs as \textbf{A}ssistants of \textbf{L}LMs). Specifically, unlike other conventional approaches, we assign different weights to entities in EMRs based on their type, which enables us to more precisely localize possible candidate diseases in KGs. Meanwhile, in order to prevent the results from relying too much on KGs, we drew inspiration from the idea of residual networks to allow LLMs to first diagnose without relying on external knowledge, and then merge the diagnosis results with the search results of the knowledge graph. Subsequently, we propose a path-based rerank algorithm to rank candidate diseases. Finally, we designed a special fill-in-the-blank style prompt template to help LLMs to better inference and error correction.
\newline \indent In summary, our contributions can be abbreviated as: (1) We raised the problem of a shortage of high-quality open-source Chinese electronic medical record data and we introduced an open-sourced Chinese EMR dataset. (2) We proposed an effective method that allows LLMs to handle information-dense and highly redundant EMRs to make correct diagnoses. (3) We conducted extensive experiments on our collected EMR dataset to demonstrate the effectiveness of our medIKAL framework.

\section{Related Work}
\subsection{Clinical Diagnosis and Prediction on EMRs}
Electronic medical records (EMRs) provide detailed medical information about patients, including symptoms, medical history, test results, and treatment records, and are widely used in patient care, clinical diagnosis, and treatment~\citep{xu2024ram-EHR}. Prior research has extensively focused on designing deep learning models for EMR data, addressing downstream tasks such as disease diagnosis and risk assessment~\citep{gao2020stagenet, xu2022counterfactual, wang2023hierarchical}.


LLMs have demonstrated impressive performance in various medical tasks, including disease diagnosis and prediction in EMRs. Researchers have explored multiple approaches: \citep{jiang2023graphcare} used LLMs and biomedical knowledge graphs to construct patient-specific knowledge graphs, processed with a Bidirectional Attention-enhanced Graph Neural Network (BAT GNN); RAM-EHR~\citep{xu2024ram-EHR} transformed multiple knowledge sources into text format, utilizing retrieval-enhanced and consistency-regularized co-training; DR.KNOWS~\citep{gao2023Dr_knows} combined a knowledge graph built with the Unified Medical Language System (UMLS) and a clinical diagnostic reasoning-based graph model for improved diagnosis accuracy and interpretability; REALM~\citep{zhu2024realm} integrated clinical notes and multivariate time-series data using LLMs and RAG technology, with an adaptive multimodal fusion network. Most studies focus on English EMR datasets like MIMIC-III~\citep{johnson2016mimic-III}, which primarily contains ICU data and may not suffice for modeling mild cases, rehabilitation, or routine treatments. Research on Chinese EMR datasets remains limited.

\subsection{Knowledge Graphs Augmented LLMs}

Knowledge graphs have advantages in dynamic, explicit, structured knowledge representation and storage, and easy addition, deletion, modification, and querying~\citep{pan2024unifying}, which has led to increasing interest among researchers in exploring the integration of knowledge graphs with large language models. One typical paradigm is to incorporate knowledge graph triplets into the training data during the training phase and obtain their embedding representations through graph neural network modules~\citep{zhang2019ernie, sun2021ernie, li2023trea, huang2024joint}.  However, LLMs often have a large-scale requirement for pre-training corpora, making it difficult and costly to find or create knowledge graphs of a matching scale~\citep{wen2023mindmap}. More importantly, combining knowledge graphs with LLMs through embedding can result in the loss of their original advantages, such as interpretability of reasoning and efficiency of knowledge updates.

In recent studies, researchers have attempted to integrate KGs with LLMs through prompts~\citep{wen2023mindmap, ICP_knowledge_seed, yang2024kgrank, wang2023keqing}. They typically identify entities in the input text and locate the corresponding triplets or subgraphs in the KG, which are then transformed into natural language~\citep{wen2023mindmap}, entity sets~\citep{ICP_knowledge_seed}, or reorganized triplets~\citep{yang2024kgrank}, etc., and concatenated with the input prompts to provide additional knowledge to LLMs. Another approach is to use an iterative strategy where LLMs act as agents to explore and reason step-by-step on the KG until it obtains sufficient knowledge or reaches the maximum number of iterations~\citep{sun2023think, jin2024graphcot}. However, this approach is more suitable for shorter questions. In scenarios with longer contexts, larger knowledge graph scales, and more complex structures, it can result in excessive interactions with LLMs and the inability to find the correct paths in the knowledge graph.

\begin{figure*}[t]
\centering
\includegraphics[scale=0.6]{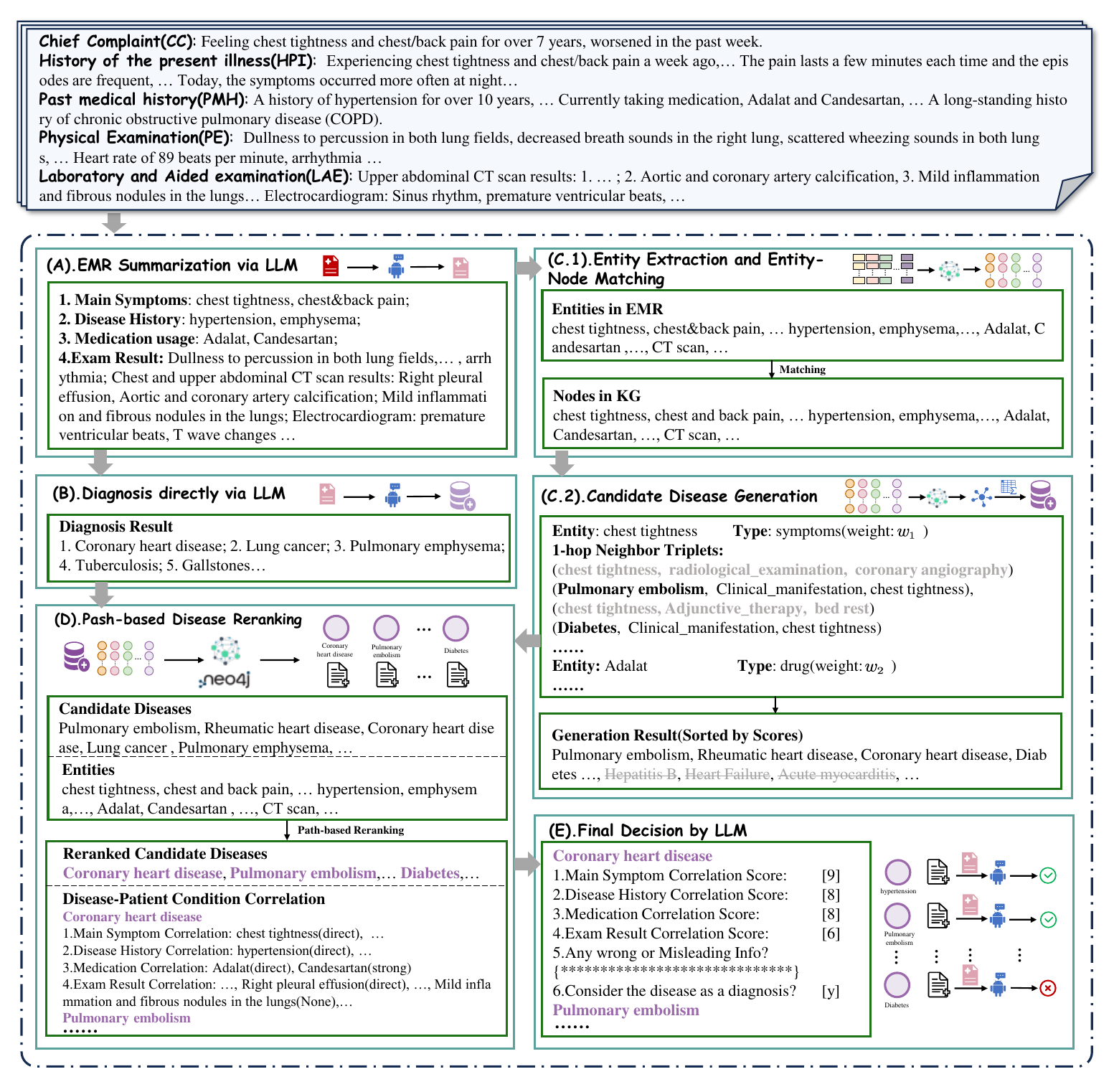}
\caption{The overall workflow of medIKAL. It contains three main modules, namely: \textbf{Module 1.}~preprocess before KG search (A, B, and C.1); \textbf{Module 2.}~Candidate Disease Localization and Reranking via KG (C.2 and D); \textbf{Module 3.}~Collaborative Reasoning for LLMs and KG (E). }
\label{workflow}
\end{figure*}

\section{Method}
\subsection{EMR Summarisation and Direct Diagnosis via LLMs}
\label{pre}

Considering that the EMRs contain a large amount of redundant information, direct use is easy to cause interference in the diagnostic process. So we first designed a series of questions to prompt the LLM to summarize the key information in the EMR, such as patient symptoms, medical history, medication usage, medical visits, etc. Detailed prompt templates are shown in Table~\ref{summary_prompt} and ~\ref{summary_exam_prompt} in Appendix~\ref{prompt_template}. This process can be represented as:
\begin{equation}
    \mathcal{M} =\mathrm{LLM}([\mathrm{Prompt_{sum}},\mathcal{M}_{orig}])
\end{equation}
where $\mathcal{M}_{orig}$ represents the original input medical record, $\mathcal{M}$ represents the medical record after decomposition and summarization, and $\mathrm{Prompt_{sum}}$ is the textual prompt.

Based on the decomposed and summarized medical record, we allow the LLM to rely on its internal knowledge for preliminary diagnosis and obtain a set of potential diseases $\mathcal{D}_{\mathrm{LLM}}$. This process can be represented as:
\begin{equation}
    \mathcal{D}_{\mathrm{LLM}} =\mathrm{LLM}([\mathrm{Prompt_{diag}},\mathcal{M}])
\end{equation}
where $\mathrm{Prompt_{diag}}$ denotes the textual instruction used to guide the LLM in performing preliminary diagnosis and providing predicted diseases~(see Table~\ref{diagnosis_prompt} in Appendix~\ref{prompt_template}).

\subsection{Candidate Disease Localization and Reranking via KG}
\subsubsection{Entity Recognition and Matching}
\label{NER_match}
Before the knowledge graph search process, we perform entity recognition on the summarized EMR $\mathcal{M}$ using a pre-trained NER model. This process can be represented as:
\begin{equation}
\begin{aligned}
&\mathcal{E}_{\mathcal{M}} = e_1, e_2, \ldots, e_{|E|} = \mathrm{NER}(\mathcal{M}) \\
\end{aligned}
\end{equation}
Where the entity set extracted from the EMR is denoted as $\mathcal{E_M}$, and $\mathrm{NER}$ denotes the pre-trained NER model.

Then for every $e_i \in \mathcal{E_M}$, we link it to the corresponding node in the knowledge graph $\mathcal{G}$ using dense retrieval methods. Specifically, given an entity $e_i \in \mathcal{E_M}$, we use an encoding model to get the embedding of $e_i$, and calculate the similarity score between $e_i$ and each entity node $u_j$ in $\mathcal{G}$'s entity node set $\mathcal{E_G}$, and the entity node with the highest similarity score is considered as a match. This process can be formulated as follows:
\begin{equation}
    \begin{aligned}
        \hat{u}_i = \arg\max_{u_j \in \mathcal{E_G}} \mathrm{sim}(\mathrm{enc}(e_i),\mathrm{enc}(u_j)), \\
    \end{aligned}
\end{equation}
Where $\mathrm{enc}$ denotes the encoding model, and $\hat{u}_i$ denotes the matched entity node. Finally, the set of matched entities is denoted as $\mathcal{E_Q}$.

\subsubsection{Candidate Disease Localization Based on Entity-Type Weights}
\label{entity_dis_location}
Most of the previous work using KG to augment LLMs has not made a strict distinction between entity types when using entities for the knowledge graph search process. However, in the EMR, different types of entities are supposed to contribute differently to the diagnosis of a disease. For example, the association between a patient's current symptoms and the disease is more direct and closer. 

So in this paper, we propose an entity type-driven method for candidate disease localization and filtering. For every entity $e_i \in \mathcal{E_Q}$, we assign a contribution weight $w_{t_i}$ according to its entity type $t_i$. Then we search for disease nodes in the 1-hop neighbors of $e_i$ in $\mathcal{G}$ and obtain the set of disease nodes $\mathcal{D}_i$, where the score of each disease in $\mathcal{D}_i$ will be increased by $w_{t_i}$. The algorithm description of the above process can be found in Algorithm~\ref{alg:can} in Appendix~\ref{algo_appendix}. 
After getting the potential disease set $\mathcal{D}_{\mathcal{G}}$ generated by the KG search process, we merge $\mathcal{D}_{\mathcal{G}}$ with the potential disease set $\mathcal{D}_{\mathrm{LLM}}$ obtained through LLMs in Section~\ref{pre}, resulting in a candidate disease set $\mathcal{D}_{can} = \mathcal{D}_{\mathrm{LLM}} \cup \mathcal{D}_{\mathcal{G}}$. Here we have drawn inspiration from the idea of residual networks~\citep{resnet}. We hope to make more use of the LLM's internal knowledge in this way, rather than relying solely on the knowledge graph for searching for correct diagnoses.

\subsubsection{Candidate Disease Reranking Based on Paths.}
\label{rerank_by_path}

\begin{figure*}[t]
\centering
\includegraphics[scale=0.3]{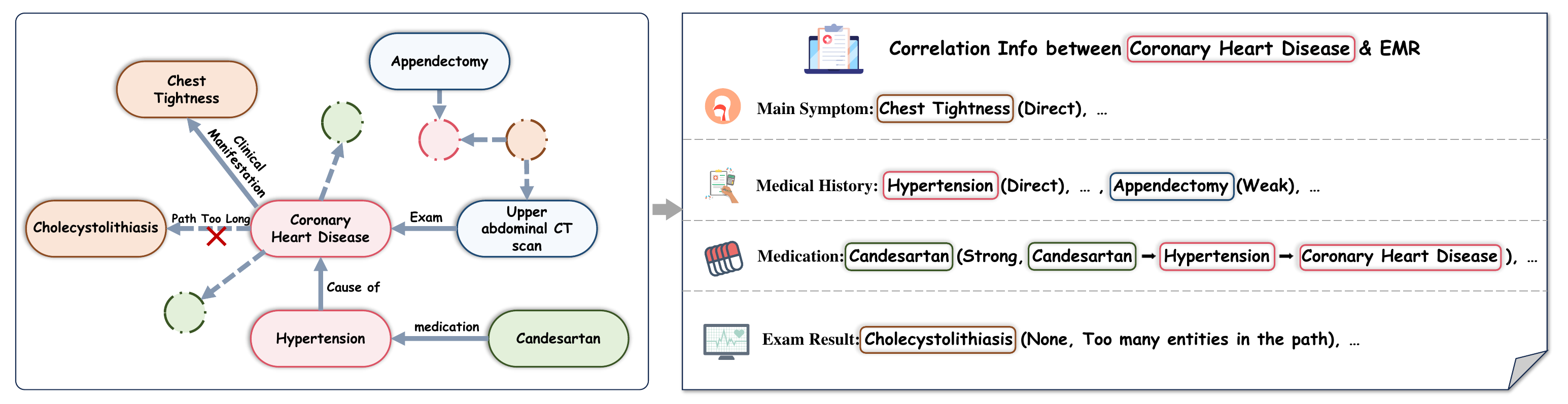}
\caption{An illustration of how to combine reranking process with the knowledge construction process.}
\label{combination}
\end{figure*}

In actual clinical diagnosis, doctors usually make a diagnosis based on a series of information such as the patient's symptoms, medical history, examination results, etc. Therefore, a correct diagnosis should be correlated with most of the patient information. In order to model this correlation, we propose a path-based reranking algorithm. Specifically, we define $\mathrm{dist}(\mathcal{D}_i, e_j)$ to denote the shortest path distance between disease $\mathcal{D}_i$ and entity $e_j \in \mathcal{E_Q}$ on $\mathcal{G}$. Diseases with closer total distances to the entity set $\mathcal{E_Q}$ are considered to have a stronger association with the patient's information, making them more likely to be the correct diagnostic results. The specific process of path-based reranking can be found in Algorithm~\ref{alg:rerank}.


\subsection{Collaborative Reasoning between LLMs and KG Knowledge}
After completing the search and reranking process based on the knowledge graph, we reconstructed the search results to provide additional contextual information for LLMs for collaborative reasoning. 
\subsubsection{Reconstruction of KG Knowledge}

EMRs are different from conventional medical QA tasks. Even though we have previously summarized them, they are still information-dense and complex-context structures, so the retrieved KG knowledge will also become extensive. If we still follow previous work and directly input triplets or knowledge chain paths as context knowledge, it would lead to overly chaotic structures that LLMs can hardly understand, which increases the possibilities of hallucination. Therefore, in this paper, we propose a way to reconstruct knowledge graph information. For each candidate disease $\mathcal{D}_i \in \mathcal{D}_{rerank}$, we classify and organize the information related to $\mathcal{D}_i$ according to several aspects like the correlations between $\mathcal{D}_i$ and the patient's main symptoms, or between $\mathcal{D}_i$ and the patient's medical history, etc. An example illustration is shown in Figure~\ref{combination}.

In this way, we transform the information of paths and entities retrieved from the knowledge graph into a semi-structured representation of knowledge, which maximizes the manifestation of the association between each candidate disease and the content of the medical record, enabling LLMs to make more intuitive judgments and analyses. Moreover, since the association between the majority of entities and diseases has already been established during the processing of Section \ref{entity_dis_location} and Section \ref{rerank_by_path}, the knowledge reconstruction process does not require re-searching $\mathcal{G}$, avoiding additional time consumption.


\subsubsection{Clinical Reasoning and Diagnosis Based on Fill-in-the-Blank Prompt Templates}
\label{FBP}
Based on the reconstructed knowledge described above, we designed a special prompt template in a fill-in-the-blank style to make the reasoning paths of LLMs more rational. We guide LLMs to quantitatively evaluate the degree of correlation between a specific disease $\mathcal{D}_i$ and the aspects mentioned above, giving a score ranging from 0 to 10 (the higher the score, the higher the degree of correlation) for each aspect, and then calculate a total score. If the total score is higher than a pre-defined threshold $\theta$, we consider the current candidate disease $\mathcal{D}_i$ as one of the final diagnostic results. Additionally, to ensure the self-consistency of LLMs, we also check the consistency between this total score and the prediction made by LLMs. If they are inconsistent, we will check the original prediction $\mathcal{D}_{\mathrm{LLM}}$ to decide whether to drop $\mathcal{D}_i$. The specific prompt template can be found in Table~\ref{analysis} in Appendix~\ref{prompt_template}. 

\section{Experiments}
\subsection{Experimental Setup}
\subsubsection{Datasets}
\textbf{CMEMR Dataset Construction}: Considering the current lack of high-quality and widely covered EMR datasets in the Chinese community, we construct a dataset CMEMR~(Chinese Multi-department Electronic Medical Records) collected from a Chinese medical website\footnote{\url{https://bingli.iiyi.com/}}. We filtered the collected electronic medical records, excluding those with existing problems or missing key information. The details of the dataset can be seen in Table~\ref{department_distribution} in Appendix~\ref{data_detail}. In order to ensure the correctness and usability of the collected medical records, we randomly sampled a batch of medical records in each department and consulted the corresponding department experts, mainly focusing on the correctness of the diagnosis results (i.e., the labels of our task). A complete data example is provided in Figue~\ref{fig:example_zh} and \ref{fig:example_en}.

In addition, to further validate our proposed method, we selected the following three datasets as supplements: (1) \textbf{CMB-Clin}~\citep{cmb}: The CMB-Clin dataset contains 74 high-quality, complex, real EMRs, each of which will contain several medical QA pairs. To be consistent with our approach, we simplify the task of this dataset to a pure disease diagnosis task. (2) \textbf{GMD}~\citep{GMD}: The GMD dataset was constructed based on EMRs. Each sample in the dataset contains a target disease along with its explicit and implicit symptom information. (3) \textbf{CMD}~\citep{CMD}: The CMD dataset is a follow-up to the GMD dataset. Its format is the same as the GMD dataset, and also sourced from EMRs. The only difference is that CMD contains a more variety of diseases and symptoms.

\begin{table*}[t]
\centering
\begin{tabular}{cl|lll|lll|lll}
\toprule
\hline
\multicolumn{2}{c|}{\multirow{2}{*}{Methods}} & \multicolumn{3}{c|}{Qwen-7b-chat} & \multicolumn{3}{c|}{Qwen-14b-chat} & \multicolumn{3}{c}{Qwen-72b-chat} \\
\cline{3-5}\cline{6-8}\cline{9-11}
\multicolumn{2}{c|}{} & R & P & F1 & R & P & F1 & R & P & F1 \\ \hline
I & Direct & 41.07  & 31.23 & 35.48 & 42.98 & 32.50 & 37.01 & 45.12 & 34.45 & 39.06 \\ \hline
\multirow{3}{*}{II} & CoT &  41.24 & 31.06 & 35.43 & 42.58 & 31.67 & 36.32 & 46.01 & 33.19 & 38.56 \\
 & ToT & 39.25 & 31.77 & 35.11 & 43.19 & 32.56 & 37.12 & 45.45 & 34.87 & 39.46 \\
 & SC-CoT & 41.99 & 31.69 & 36.12 & 42.34 & 32.90 & 37.40 & 45.49 & 34.59 & 39.29 \\ \hline
\multicolumn{1}{l}{\multirow{6}{*}{III}} & MindMap & 41.42 & 32.30 & 36.29 & 43.59 & \textbf{33.81} & 38.08 & 45.14 & 35.62 & 39.81 \\
\multicolumn{1}{l}{} & KG-Rank & 39.13 & 28.61 & 33.05 & 41.34 & 31.45 & 35.72 & 44.79 & 32.95 & 37.96 \\
\multicolumn{1}{l}{} & ICP & 40.13 & 30.67 & 34.76 & 41.58 & 30.23 & 35.00 &  44.00& 32.38 & 37.30 \\
\multicolumn{1}{l}{} & HyKGE & 42.05 & 32.42 & 36.61 & 43.76 & 33.45 & 37.91 & 45.91 & 34.30 & 39.26 \\
\multicolumn{1}{l}{} & ToG & 38.78 & 26.94 & 31.79 & 39.09 & 27.31 & 32.15 & 40.39 & 27.81 & 32.93 \\
\multicolumn{1}{l}{} & Graph-CoT & 35.90 & 24.01 & 28.77 & 38.67 & 25.11 & 30.44 & 39.68 & 27.48 & 32.47 \\
\hline
\multicolumn{1}{l}{} & Ours & \textbf{42.16} & \textbf{32.86} & \textbf{36.93}  & \textbf{43.96}  & 33.65  & \textbf{38.12}  &\textbf{46.43}  & \textbf{35.72}  & \textbf{40.37}  \\
\hline
\bottomrule
\end{tabular}
\caption{Experimental results on CMEMR dataset with different scale of backbone models. The best results are highlighted in bold.}
\label{main_EMR}
\end{table*}

\subsubsection{Baselines}
We compared our proposed medIKAL with three series of baseline methods: LLM-only, LLM$\oplus$KG, and LLM$\otimes$KG~\citep{sun2023think}. Details of the baseline methods are provided in Appendix \ref{baseline_details}.

\noindent\textbf{LLM-only:}~ They do not rely on external knowledge and only use the LLMs' internal knowledge for reasoning, including CoT~\citep{wei2022chain}, ToT~\citep{yao2024tree}, and Sc-CoT~\citep{wang2022self}).

\noindent\textbf{LLM$\oplus$KG:} ~We selected four representative works, namely MindMap~\citep{wen2023mindmap},  ICP~\citep{ICP_knowledge_seed}, HyKGE~\citep{HyKGE}, , and KG-rank~\citep{yang2024kgrank}, all of which are aimed at medical question-answering and reasoning tasks, so we believe they are highly relevant to our work in this paper.


\noindent\textbf{LLM$\otimes$KG:}~ This is the concept proposed by~\citep{sun2023think}. It enables LLMs to participate in the search and reasoning process on KGs, check whether the current knowledge is sufficient to answer the question, and make decisions for the subsequent search process iteratively. We selected ToG~\citep{sun2023think} and Graph Chain-of-Thought~\citep{jin2024graphcot} as baselines.

\subsubsection{Evaluation metric}
To enhance the scientific rigor and effectiveness of the evaluation, particularly in identifying disease diagnoses, following \citep{fan2024ai-hospital}, we adopted the International Classification of Diseases (ICD-10) \citep{percy1990ICD} as the authoritative source and link standardized disease terminologies with natural language based diagnostic results. Initially, we extract disease entities from the diagnostic results and the label in the EMR. Then we implement a fuzzy matching process with a predefined threshold of 0.5 to link these disease entities with ICD-10 terminology, building two normalized disease sets $S_{\mathcal{\hat{D}}}$ and $S_{\mathcal{R}}$. Finally we use these two sets to calculate the Precision, Recall and F1-score metrics. More details are shown in Appendix~\ref{metrics}.


\begin{table*}[htp]
\centering
\begin{tabular}{cl|lll|lll|lll}
\toprule
\hline
\multicolumn{2}{c|}{\multirow{2}{*}{Methods}} & \multicolumn{3}{c|}{CMB-Clin} & \multicolumn{3}{c|}{GMD} & \multicolumn{3}{c}{CMD} \\
\cline{3-5}\cline{6-8}\cline{9-11}
\multicolumn{2}{c|}{} & R & P & F1 & R & P & F1 & R & P & F1 \\ \hline
I & Direct & 40.35 & 26.77  &32.18  & 42.01 & 21.03 & 28.02 & 50.26 & 25.11 & 33.48 \\ \hline
\multirow{3}{*}{II} & CoT & 40.66  & 27.23  & 32.62 & 42.44 & 21.30 & 28.36 & 51.02 & 25.49 &  33.99\\
 & ToT & 39.94 & 25.90 & 31.42 & 41.68 & 20.80 & 27.75 & 49.39 & 24.48 & 32.73 \\
 & SC-CoT & 41.10 & 26.31 & 32.08 & \textbf{42.73} & 21.37 & \textbf{28.49} & 51.14 & 25.57 & 34.09 \\ \hline
\multicolumn{1}{l}{\multirow{6}{*}{III}} & MindMap & 39.26  & \textbf{29.24}  & 33.51 & 41.44 & 21.18 & 28.03 & 49.75 & 25.62 & 33.82 \\
\multicolumn{1}{l}{} & KG-Rank & 41.70  & 27.12  & 32.86 & 38.16 & 19.54 & 25.84 & 47.91 & 23.92 & 31.90  \\
\multicolumn{1}{l}{} & ICP & 40.27  & 25.54 & 31.25  & 39.38 & 19.63 & 26.20 & 46.26 & 23.15 & 30.85 \\
\multicolumn{1}{l}{} & HyKGE & 41.53 & 28.21 & \textbf{33.59} & 40.33 & 21.36 & 27.92 & 48.67  & 24.35 & 32.45 \\
\multicolumn{1}{l}{} & ToG & 35.41 & 19.18 & 24.88  & 41.76 & 20.85 & 27.81 & 50.73 & 25.24 & 33.70 \\
\multicolumn{1}{l}{} & Graph-CoT & 36.35 & 20.66 & 26.07 & 38.13 & 19.06 & 25.54 & 49.07 & 24.51 & 32.69 \\
\hline
\multicolumn{1}{l}{} & Ours & \textbf{41.89}  & 27.68  &33.33  & 42.37 & \textbf{21.43} & 28.46  & \textbf{51.26} & \textbf{25.74} & \textbf{34.27} \\
\hline
\bottomrule
\end{tabular}
\caption{Experimental results on CMB-Clin, GMD, and CMD datasets using Qwen-7B-chat. The best results are highlighted in bold.}
\label{main_tiny}
\end{table*}

\subsubsection{Implementation Details}
\label{implementation_detail}
For the backbone model, we choose Qwen models with different parameter scales~([7B, 14B, 72B]). In all experiments, we set $do\_sample$ to false for consistent responses. For the knowledge graph, we choose the CPubMed-KG. For the NER model mentioned in section~\ref{NER_match}, we choose the RaNER~\citep{RaNER} model released by Tongyi-Laboratory. For the Entity-node matching process in section~\ref{NER_match}, we choose the CoROM~\citep{corom} model as our embedding model. More implementation details are listed in Appendix~\ref{overall_detail}.


\subsection{Experimental Results}
\subsubsection{Overall Performance}
The main experimental results on CMEMR dataset are shown in Table~\ref{main_EMR}. From the results, we can draw the following analysis: 

\noindent\textbf{(1)} Our method significantly outperforms other baselines using \textbf{LLM$\oplus$KG} paradigm on CMEMR dataset, which demonstrated the effectiveness of our method on EMR-diagnosis task. 

\noindent\textbf{(2}) The methods using \textbf{LLM$\otimes$KG} (i.e., ToG~\citep{sun2023think} and Graph-CoT~\citep{jin2024graphcot}) perform poorly on EMR-diagnosis Tasks, since they are designed for short multi-hop QA task. The iteration steps and the complexity of beam search increase greatly as the amount of context and the size of KG increase, which makes it easily reach the upper limit of the number of iterative steps without collecting enough information, or exceeding the input length limit of LLMs.

\noindent\textbf{(3)} As we expected, the performance of medIKAL improves with the scale of backbone models due to the increase of models' reasoning and instruction-following ability. Considering the plug-and-play and train-free nature of our method, it can be flexibly deployed to backbone models of different sizes depending on the needs of different scenarios.

We also tested our method on three additional datasets and the experimental results are shown in Table~\ref{main_tiny}. Our method performs stably on the CMB-Clin dataset, whose data format is also standard EMRs. On the GMD and CMD datasets, there is a slight degradation in the performance of our method. This is because although GMDs and CMDs are also constructed using EMRs, they contain too little patient information (only symptoms), which can easily localize to other related diseases on the knowledge graph leading to errors.

\subsubsection{In-depth Analysis}
\begin{table}[h]
\centering
\begin{tabular}{l|lll}
\toprule
\hline
Methods & R & P & F1 \\ \hline
Relevant Entities & 39.22 & 28.74 & 33.17 \\
Natural Language & 39.88 & 28.92 &  33.52\\
Relevant Triples & 40.26 & 29.61 & 34.12 \\
Reasoning Chains & 40.97 & 31.16 & 35.39 \\
MindMap & 41.10 & 31.41 & 35.60\\
\hline
FBP(ours) & \textbf{42.16} & \textbf{32.86} & \textbf{36.93}\\
\hline
\bottomrule
\end{tabular}
\caption{Performances of medIKAL using different knowledge graph-augmented prompt templates on CMEMR dataset. Note that we kept all the rest parts of the medIKAL and only replaced the final “fill-in-the-blanks” prompts (FBP) with other methods to conduct this experiment.}
\label{knowledge_representation}
\end{table}


\noindent\textbf{How do different knowledge graph augmented prompts affect medIKAL's performance?}\indent  In order to verify our proposed special prompt template's superiority, we compare it with several knowledge graph-augmented prompt templates, including entities~\citep{ICP_knowledge_seed}, relevant triplets~\citep{yang2024kgrank}, natural language, reasoning chains~\citep{HyKGE}, and mindmap~\citep{wen2023mindmap}. The experimental results are shown in Table~\ref{knowledge_representation}. According to the results, using relevant entities is very ineffective as it does not utilize the relational information contained in the knowledge graph at all. For the reasoning chains and mindmap, due to the information-intensive nature of EMR data, they can easily form overly large and complex-structure prompt contexts, making it difficult for LLMs (especially models with small parameters) to reason.

\begin{figure}[h]
\centering
\includegraphics[scale=0.40]{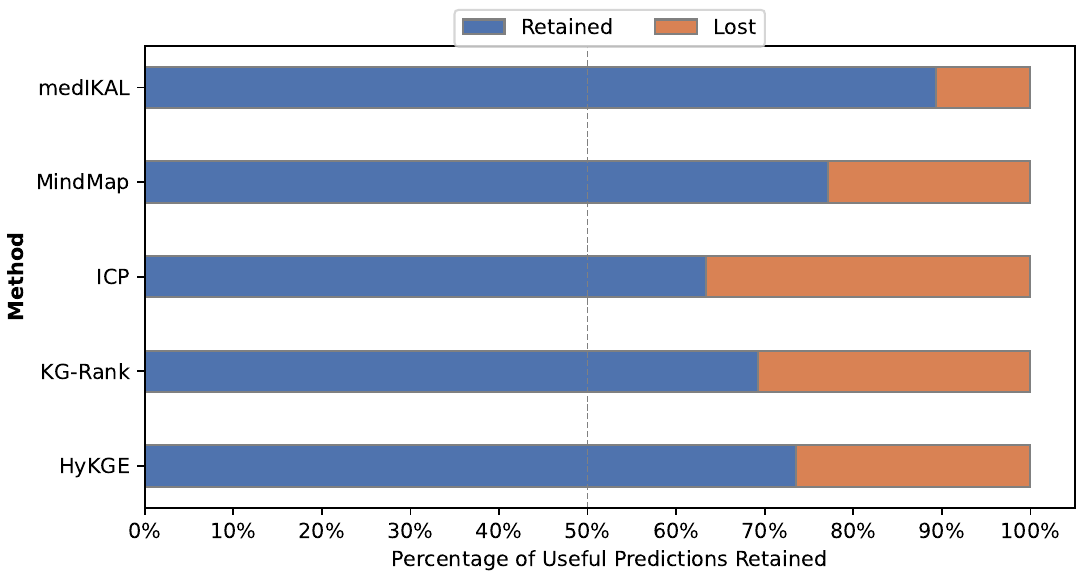}
\caption{Evaluation results for medIKAL and other baseline methods' capabilities of utilizing LLM's internal knowledge. "Retained" denotes that the useful diagnoses from LLM's original predictions are kept as final results, and "Lost" denotes the opposite. }
\label{retain}
\end{figure}

\noindent \textbf{Does medIKAL integrate KG and LLMs better compared with other baselines?} \indent The problem with most of the existing work based on knowledge graphs is that LLMs can be overly dependent on the information obtained from KG and fail to use their own knowledge. Therefore, we counted the proportion of useful predictions in the original predictions of LLMs retained by medIKAL and other baseline methods. From the experimental results in Figure~\ref{retain}, medIKAL is able to minimize LLM's over-reliance on KG's knowledge and retains the majority of useful predictions compared to other baselines. 

\subsubsection{Ablation Study}
\begin{table}[h]
\centering
\begin{tabular}{l|lll}
\toprule
\hline
Method & R & P & F1 \\ \hline
medIKAL & \textbf{42.16} & \textbf{32.86} & \textbf{36.93}\\
\hline
\emph{w/o} SUM & 41.56 & 32.37 & 36.39 \\
\emph{w/o} ETW & 41.19 & 29.88 & 34.63 \\
\emph{w/o} PR & 41.91 & 32.44 & 36.57 \\
\emph{w/o} RI & 40.16 & 30.32 &  34.55\\
\hline
\bottomrule
\end{tabular}
\caption{Ablation study results on CMEMR dataset. \emph{w/o} indicates removal of the corresponding module. "SUM" denotes "summarization". "ETW" denotes "\textbf{E}ntity \textbf{T}ype \textbf{W}eight". "PR" denotes "\textbf{P}ath-based \textbf{R}eranking". "RI" denotes "\textbf{R}esnet-like \textbf{I}ntegration".}
\label{ablation}
\end{table}
We conduct the following ablation studies to demonstrate the importance of different modules in medIKAL.  

\noindent\textbf{(a)}.\textbf{\emph{w/o} SUM}~(summarization): Remove the summarization step when pre-processing medical records and i nstead use the raw content directly. \\
\textbf{(b)}.\textbf{\emph{w/o} ETW}~(Entity-Type Weight): Remove the entity-type weight when performing entity-based candidate disease searches, with all entities contributing equal weights. \\
\textbf{(c)}.\textbf{\emph{w/o} PR}~(Path-based Reranking): Remove the reranking process for candidate diseases.\\
\textbf{(d)}.\textbf{\emph{w/o} RI}~(Resnet-like Integration): Do not integrate the LLM's direct diagnosis result into the candidate disease.

The results in Table~\ref{ablation} show that both removing the “SUM” module and the “ETW” settings can seriously interfere with the performance, as the former leads to the introduction of a lot of redundant information in the original EMRs, while the latter leads to unimportant entities overly influencing the results. Removing the “RI” module would result in results that are entirely dependent on the KG search process, while the internal knowledge of LLMs is almost completely unused, thus causing a severe performance decrease. 

%
\subsubsection{Case Study}
\label{case}
We show representative case studies in Figure~\ref{case_study} in the Appendix to demonstrate the effectiveness of our proposed medIKAL. From Figure~\ref{case_study}, we can find that medIKAL can not only complement (Figure~\ref{case_study}-(a)) and correct (Figure~\ref{case_study}-(d)) the predictions of LLMs using KG, but also effectively guide LLMs to analyze and reason (Figure~\ref{case_study}-(b)). Besides, the cross-validation approach through quantitative assessment and model judgment can also effectively improve the fault tolerance for LLMs' hallucination(Figure~\ref{case_study}-(c)).

\subsubsection{Error Analysis}
\label{error_study}
\textcolor{black}{We have analyzed the errors of our medIKAL framework on the CMEMR dataset. Through a systematic manual review, these errors are mainly categorized into three distinct classes, which are detailed in Table~\ref{error_analysis} in the Appendix.}

\noindent\textbf{Overlapping Disease Characteristics} \indent
\textcolor{black}{This type of error can be divided into two cases:  
(1) For distinct diseases with significant feature overlap (e.g., symptoms, affected sites, diagnostic methods, medications), LLMs may misdiagnose during preliminary diagnosis based on patient information. Similarly, feature overlap during the KG-retrieval stage often leads to diagnostic failures.  
(2) For a disease $d$ and its subtypes (e.g., "anemia" and "iron-deficiency anemia"), nodes related to subtype diseases in the KG often have similar or stronger associations with the parent disease $d$. Consequently, after retrieval and re-ranking, the parent disease tends to rank higher.}

\noindent\textbf{Misunderstanding of Examination Indicators}\indent  
\textcolor{black}{Specific examination indicators, such as numerical values or symbols, are challenging to effectively map to knowledge graph nodes (e.g., "EF 27\%" may be mapped to multiple nodes like "EF value elevated," "EF value reduced," or "EF value normal"). This ambiguity makes accurate mapping difficult. Additionally, LLMs struggle with recognizing and using such indicators, often generating hallucinations when interpreting numerical ranges.}

\noindent\textbf{Gap Between LLM Reasoning and Physician Diagnostic Labeling}\indent  
\textcolor{black}{This error stems from dataset structural issues. For example, some physicians include past diseases in current diagnosis records, even when irrelevant to the present complaint, but our framework does not account for this. Moreover, abbreviations used by doctors for convenience may not be accurately interpreted by LLMs, leading to the exclusion of correct candidate diseases.}

\section{Conclusion}
In this paper, we proposed medIKAL, a framework that seamlessly integrates LLMs with knowledge graphs to enhance clinical diagnosis on EMRs, with its key innovation being the weighted importance assignment to medical entities and a resnet-like integration approach. Experimental results showed that medIKAL significantly outperforms baselines, demonstrating its potential to improve diagnostic accuracy and efficiency in real-world clinical settings. Our medIKAL has offered a promising direction for AI-assisted clinical diagnosis, paving the way for more advanced healthcare applications.
\section*{Limitations}
\noindent\textbf{The limitations of collected CMEMR dataset.} 
Although we have meticulously examined, desensitized, and verified the CMEMR dataset with medical experts, occasionally, the quality of the medical records may still fall short in actual experiments. Additionally, due to the limited sources of data, our medical record dataset exhibits an uneven distribution across departments.

\noindent\textbf{The limitations of proposed medIKAL framework.} \indent Although medIKAL has demonstrated its effectiveness and great potential in the healthcare field, it still has some limitations. Firstly, while it is not strictly limited to EMR format inputs, it requires a high amount of information from the input data samples. When the input data information is sparse, the improvement in model reasoning performance by medIKAL decreases, and there is also an increased risk of hallucinations. Furthermore, medIKAL is unable to fully utilize special types of medical examination indicators (e.g., numerical or symbolic types). Addressing this issue is a key problem that needs to be solved in our future work.

\section*{Ethical Consideration}
\textcolor{black}{In our study on the application of LLMs in clinical disease diagnosis, ethical considerations are of paramount importance. We acknowledge the potential impacts of our work and have taken measures to address these concerns. To mitigate risks such as privacy breaches and the exposure of personal information, we have thoroughly reviewed and de-identified the data. Regarding copyright concerns, we plan to split the dataset by medical departments and store each EMR sample using its specific ID. Researchers will be able to access the full EMR data conveniently via the corresponding ID and the URL of the original source website.}

\section*{Acknowledgments}
\textcolor{black}{This work was supported in part by the National Key Research and Development Program of China (No.2021YFF1201200), the Science and Technology Major Project of Changsha (No.kh2402004). This work was carried out in part using computing resources at the High-Performance Computing Center of Central South University.}

\normalem
\bibliography{ref}

\appendix

\section{Detailed Information of the CMEMR dataset}
Specific information on the CMEMR dataset is shown in Table~\ref{department_distribution}.
\label{data_detail}
\begin{table}[h]
    \centering
    \begin{tabular}{lll}
    \toprule
    \hline
        Department & Num & Avg Len \\ 
    \hline
        Gynaecology & 411 & 627.46\\ 
        Otolaryngology  & 212  & 967.99\\
        Obstetrics\&Gynecology & 1316 & 489.15\\
        Nursing & 52 & 584.88\\
        Emergency & 87 & 552.96\\
        Psychiatry & 127 & 867.66\\
        Rehabilitation & 284 & 631.13\\
        Dentistry & 130 & 342.56\\
        Anesthesiology & 232 & 634.25\\
        Internal Medicine & 3590 & 528.72\\
        Dermatology & 286 & 518.08\\
        Neurosurgery & 3152  & 531.82\\
        Ophthalmologic & 100 & 453.24\\
        Oncology & 471 & 855.66\\
    \hline
        Total & 10450 & 558.60\\ 
    \hline
    \bottomrule
    \end{tabular}
    \caption{Departments distribution of the collected EMRs. "Num" denotes the total number of EMRs of the department. "Avg Len" denotes the average number of words per record.}
    \label{department_distribution}
\end{table}

Following the way of \citep{yan2024clinicallab}, we select a representative sample from CMEMR as an illustrative example, and present both the original Chinese version (Figure~\ref{fig:example_zh}) and the corresponding English translation (Figure~\ref{fig:example_en}).


\section{Algorithms for medIKAL}
\label{algo_appendix}

We summarize the comprehensive algorithmic procedure of Entity Type-driven Candidate Disease Localization and Filtering and Path-based
Reranking, as shown in Algorithm~\ref{alg:can} and~\ref{alg:rerank}.

\begin{algorithm}
\caption{Entity Type-driven Candidate Disease Localization and Filtering}
\label{alg:can}
\begin{algorithmic}[1]
\REQUIRE Entity Set $\mathcal{E_Q}$, Knowledge graph $\mathcal{G}$, Number of candidate diseases $topm$
\ENSURE Candidate disease set $\mathcal{D}_{can}$

\STATE Initialize the set of diseases $\mathcal{D} \leftarrow \emptyset$

\FOR{each entity $e_i \in \mathcal{E_Q}$}
    \STATE Assign a contribution weight $w_{t_i}$ according to its entity type $t_i$
    \STATE Obtain 1-hop neighbor triplets in $\mathcal{G}$ to locate relevant diseases $\mathcal{D}_i = \{d_{i1}, d_{i2}, \ldots, d_{in}\}$  
    \FOR{each disease $d_{ij} \in \mathcal{D}_i$}
        \IF{$d_{ij} \in \mathcal{D}$}
            \STATE Add $w_{t_i}$ to the score of $d_{ij}$
        \ELSE
            \STATE Add $d_{ij}$ to $\mathcal{D}$ with an initial score $w_{t_i}$
        \ENDIF
    \ENDFOR
\ENDFOR

\STATE Sort the diseases in $\mathcal{D}$ in descending order based on their scores
\STATE Select the $topm$ diseases to form $\mathcal{D}_{\mathcal{G}}$

\STATE Merge $\mathcal{D}_{\mathcal{G}}$ with $\mathcal{D}_{\mathrm{LLM}}$ to form $\mathcal{D}_{can} \leftarrow \mathcal{D}_{\mathrm{LLM}} \cup \mathcal{D}_{\mathcal{G}}$

\RETURN $\mathcal{D}_{can}$

\end{algorithmic}
\end{algorithm}

\begin{algorithm}
\caption{Candidate Disease Reranking Based on Paths}
\label{alg:rerank}
\begin{algorithmic}[1]
\REQUIRE Subgraph $\mathcal{G}_s = (V, E)$, Set of candidate diseases $\mathcal{D}_{can}$, Set of entities $\mathcal{E_Q}$, Number of reranked candidate diseases $topn$
\ENSURE Reranked candidate diseases $\mathcal{D}_{rerank}$

\STATE Initialize an empty list $\text{scores}$

\FOR{each disease $\mathcal{D}_i \in \mathcal{D}_{can}$}
    \STATE Initialize $\text{score} \leftarrow 0$
    \FOR{each entity $e_j \in \mathcal{E_G}$}
        \STATE Compute the shortest path $\text{dist}(\mathcal{D}_i, e_j)$
        \IF{$\text{dist}(\mathcal{D}_i, e_j) = \infty$}
            \STATE $\text{score} \leftarrow \text{score} + 0$
        \ELSE
            \STATE $\text{score} \leftarrow \text{score} + \frac{1}{\text{dist}(\mathcal{D}_i, e_j)}$
        \ENDIF
    \ENDFOR
    \STATE Append $(\mathcal{D}_i, \text{score})$ to $\text{scores}$
\ENDFOR

\STATE Sort $\text{scores}$ by the second element (score) in descending order
\STATE $\mathcal{D}_{rerank} \leftarrow$ Select the first $topn$ elements from $\text{scores}$

\RETURN $\mathcal{D}_{rerank}$
\end{algorithmic}
\end{algorithm}

\section{Detailed Setting-ups for Different Modules in medIKAL Workflow}
\label{overall_detail}

\subsection{Details of the NER Model}
\label{NER_model}
The RaNER~\citep{RaNER} model we use in this paper is released by Tongyi-Laboratory, which is trained on the CMeEE dataset~\citep{zhang2022cblue}. RaNER adopts the Transformer-CRF model, using StructBERT as the pre-trained model base, integrating the relevant sentences recalled by external tools as additional context, and employing Multi-view Training for training. It can recognize a total of 9 types of entities, including body (bod), department (dep), disease (dis), drugs (dru), medical equipment (equ), medical examination items (ite), microorganisms (mic), medical procedures (pro), and clinical symptoms (sym). 
\subsection{Details of Retrieval Method}
\label{embedding_model}
\begin{table}[t]
\centering
\begin{tabular}{l|lll}
\toprule
\hline
Retriever & R & P & F1 \\ \hline
bm25 & 40.37 & 29.86 & 34.32 \\
tf-idf & 40.25 & 29.68 &  34.16\\
\hline
m3e & 41.95 & 32.63 & 36.70 \\
all-mpnet & 42.01 & 32.75 & 36.80 \\
bge & \textbf{42.20} & 32.81 & 36.91 \\
corom & 42.16 & \textbf{32.86} & \textbf{36.93}  \\
\hline
bge + bm25 & 41.62 & 30.57 & 35.24   \\
corom + bm25 & 41.75 & 30.46 & 35.22 \\
\hline
\bottomrule
\end{tabular}
\caption{Performances of medIKAL using different retrieval methods during entity-node matching on CMEMR dataset.}
\label{retrieval}
\end{table}
In the entity-node matching process mentioned in section~\ref{NER_match}, we used a dense retrieval method to link EMR's entities to KG's nodes. In order to better explore the appropriate retrieval method, we implemented three types of retrieval methods based on the retriv library\footnote{\url{https://github.com/AmenRa/retriv}}: sparse retrieval, dense retrieval, and hybrid retrieval. 
\begin{itemize}
\item Sparse Retrieval: We evaluated two representative methods, namely bm25 and tf-idf.
\item Dense Retrieval: We evaluated several representative embedding models, namely m3e-large~\citep{Moka}, all-mpnet-base-v2, bge-large-zh-v1.5, and CoROM.
\item Hybrid Retrieval: We evaluated two combinations: "bge + bm25" and "corom + bm25".
\end{itemize}
The results are shown in Table~\ref{retrieval}. As we expected, the effect of dense retrieval is better than that of sparse retrieval and hybrid retrieval, because when the entity to be retrieved contains a large number of Chinese characters, sparse retrieval methods are prone to mismatching due to the lack of consideration of word order and semantics. According to the results, we choose the CoROM model as embedding model of the dense retrieval process.

The CoROM Chinese-medical text representation model we use in this paper is also released by Tongyi-Laboratory. It employs the classic dual-encoder text representation model and is trained on medical domain data with Multi-CPR~\citep{corom}. The training process is divided into two stages – in the first stage, negative sample data is randomly sampled from the official document set, and in the second stage, difficult negative samples are mined via Dense Retrieval to augment the training data for retraining. 

\subsection{Details of Other HyperParameters}
For the threshold $\theta$ mentioned in Section \ref{FBP}, we set it to 60\% of the total score. This parameter generally has a minor impact on performance, as for most samples, the knowledge retrieved from KGs and the patient information extracted from EMRs are sufficient for LLMs to make a confident judgment (i.e., the evaluation score is either close to full marks or close to zero), and variations in the $\theta$ value do not significantly affect the final result. But for a small number of samples with higher uncertainty, LLMs tend to provide an evaluation score close to 50\% of the total score. So after experimental comparison, we finally set $\theta$ to 60\% of the total score.

\indent To explore the influence of the number of candidate diseases \emph{Top-k} on medIKAL’s performance, we conduct experiments under settings with \emph{Top-k} ranging in [1, 2, 3, 5]. The results are shown in Table~\ref{topk}. According to the results, the Recall gradually decreases with the increase of \emph{Top-k}, while the Precision increases. When the \emph{Top-k} is set very large or very small, although it can get a higher recall or precision rate accordingly, but from the practical clinical application scenario, too large or too small \emph{Top-k} is not conducive to assisting doctors in clinical diagnosis and decision-making. Therefore, in this paper we set \emph{Top-k} to 3 on CMEMR dataset, and 2 on CMB-Clin, GMD and CMD datasets.

\begin{table}[!h]
\centering
\begin{tabular}{c|lll}
\toprule
\hline
\emph{Top-k} & R & P & F1 \\ \hline
1 & 27.27 & 56.74 & 36.83 \\
2 & 34.15 & 41.21 & 37.34 \\
3 & 42.16 & 32.86 & 36.93 \\
5 & 49.42 & 24.27 & 32.55 \\
10 & 60.85 & 13.92 & 22.74\\
\hline
\bottomrule
\end{tabular}
\caption{Performances of medIKAL with different numbers of candidate diseases~(denoted as \emph{Top-k}) on CMEMR dataset.}
\label{topk}
\end{table}

\subsection{Detailed Settings about Knowledge Graph}
\label{kg_appendix}

The knowledge graph we use in this paper is CPubMedKG-v1(Large-scale Chinese Open Medical Knowledge Graph)\footnote{\url{https://cpubmed.openi.org.cn/graph/wiki}} developed by Harbin Institute of Technology (Shenzhen). It is currently the largest fully open Chinese medical knowledge graph in China. The knowledge is derived from over 2 million high-quality Chinese core medical journals under the umbrella of the Chinese Medical Association. It is regularly updated and conforms to mainstream Chinese medical standards in terms of entity and relationship specifications. The sources of entities and relationships are clearly defined, traceable, and easily distinguishable. The graph contains a total of 4,383,910 disease-centered triples. It includes 523,052 disease entities, 188,667 drug entities, 145,908 symptom entities, and a total of 1,728,670 entities. There are more than 40 types of relationships covering drug treatment, complications, laboratory tests, indications, risk factors, affected populations, mortality rates, and more. The total number of structured knowledge triples reaches 3.9 million.

\begin{table}[h]
\centering
\begin{tabular}{p{0.4\textwidth}}
\toprule
\hline
(1) Example:\\
\textbf{[Diagnosis]}: History of vitamin D overdose. Early elevation of blood calcium > 3 mmol/L (12 mg/dl), strong positive urinary calcium (Sulkowitch reaction), routine urinalysis shows positive urinary proteins, and in severe cases, red blood cells, leukocytes, and tubular patterns are seen.\\
\hline
(2) Example:\\
\textbf{[Diagnostic Evidence]}: 1.history of prior radiotherapy for esophageal cancer, long history of hypertension, history of smoking. 2.left limb weakness for 1 day. 3.Examination revealed hypertension, decreased muscle strength of the left limb, and decreased tenderness. 4.Ancillary tests showed immediate elevated blood glucose, ECG T-wave abnormality, cervical vascular ultrasound and cranial CT and MRI suggestive of cerebral infarction.\\
\hline
\bottomrule
\end{tabular}
\caption{(1).A specific example of paragraphs related to diagnosis from the medical textbooks provided by~\citep{medqa}. (2).A specific example of diagnostic evidences in our collected EMRs.}
\label{example_textbook_evidence}
\end{table}

For the entity-type weights, we obtain the entity-type weight allocation scores through the following two methods:
\begin{itemize}
    \item We extract paragraphs related to diagnosis from the medical textbooks provided by \citep{medqa}. A specific example can be found in Table~\ref{example_textbook_evidence}-(1).
    \item We selected 500 medical records with detailed diagnostic evidence from our collection and collected all diagnostic evidence. These samples will be excluded in subsequent evaluation phases. A specific example can be found in Table~\ref{example_textbook_evidence}-(2).
\end{itemize}
We calculate the entity-type proportions of all the segments above, obtaining initial entity-type weights. The setting of our experiments can be found in Table~\ref{Entity_Type_weight_setting}. It is important to note that entity-type weights are not fixed and can be adjusted according to different tasks, which is also the advantage of the method we propose.


For the shortest path algorithm in path-based reranking, we use the GraphDataScience~\footnote{\url{https://neo4j.com/product/graph-data-science/}} library to implement it.


\begin{table}[h]
    \centering
    \begin{tabular}{ll}
    \toprule
    \hline
        Type & Weight \\ 
    \hline
        dis & .1638 \\ 
        pro & .0043 \\
        sym & .6297 \\
        dru & .1391 \\
        bod & .0212 \\
        ite & .0372\\
        equ & .0029 \\
        mic & .0009 \\
        dep & .0004 \\
    \hline
    \bottomrule
    \end{tabular}
    \caption{Entity-type weight settings in our experiments.}
    \label{Entity_Type_weight_setting}
\end{table}

\section{Details of Representative Baseline Methods}
\label{baseline_details}
\textcolor{black}{In this paper, in addition to directly using LLMs, we compare our framework with two important paradigms, namely LLM$\oplus$KG and LLM$\otimes$KG~\citep{sun2023think}. Below, we provide detailed explanations of the representative baseline methods corresponding to these two paradigms.}

\subsection{LLM$\oplus$KG Baselines}

\noindent\textbf{MindMap}~\citep{wen2023mindmap}: \textcolor{black}{The process begins by identifying key entities in the question and retrieving the knowledge graph to form evidence subgraphs. The LLM then aggregates these into a reasoning graph and generates an answer, presenting its reasoning as a mind map.}

\noindent\textbf{HyKGE}~\citep{HyKGE}: \textcolor{black}{The method follows a multi-stage pipeline: it uses LLMs' zero-shot capabilities to expand queries and identify anchor entities, retrieves reasoning chains (path, common ancestor, and co-occurrence), and re-ranks them for alignment with the query. The filtered knowledge is then combined with the query, and LLMs generate the final answer.}

\subsection{LLM$\otimes$KG Baselines}

\noindent\textbf{ToG}~\citep{sun2023think}: \textcolor{black}{In this framework, the LLM acts as an agent, using beam search to explore reasoning paths on the KG until enough information is gathered or the search depth limit is reached. ToG involves three stages: initialization, exploration, and reasoning. The LLM identifies initial entities, expands paths through search and pruning, and evaluates if the path suffices to generate an answer, repeating exploration if necessary.}

\noindent\textbf{Graph-CoT}~\citep{jin2024graphcot}: \textcolor{black}{The method simulates human thought by breaking complex graph reasoning into iterative steps: LLM reasoning to identify needed information, LLM-graph interaction to generate operations like node lookup, and graph execution to perform these operations and return results. This cycle repeats until the LLM reaches the final answer.}

\section{Evaluation Metrics Calculation}
\label{metrics}
Firstly, for the disease entities in the diagnosis results $\mathcal{\hat{D}}$ and the reference diagnosis results $\mathcal{R}$ in the medical records, we used a fuzzy matching process (with a predefined threshold of 0.5) to associate these disease entities with ICD-10 terms, thus mapping $\mathcal{\hat{D}}$ and $\mathcal{R}$ to two standardized disease sets $S_{\mathcal{\hat{D}}}$ and $S_{\mathcal{R}}$ respectively. We then define:
\textbf{True Positives (TP):}~ The number of disease entities in the predicted result $S_{\mathcal{\hat{D}}}$ that correspond correctly with the reference diagnosis $S_{\mathcal{R}}$.

\textbf{False Positives (FP):}~ The number of disease entities that appear in the predicted result $S_{\mathcal{\hat{D}}}$ but do not match correctly with the reference diagnosis $S_{\mathcal{R}}$.

\textbf{False Negatives (FN):}~ The number of disease entities in the reference diagnosis $S_{\mathcal{R}}$ that do not appear in the predicted result $S_{\mathcal{\hat{D}}}$.

Based on the above statistical values, we calculate the following evaluation metrics:
\begin{align}
&\mathrm{Recall~(R)}:R=\frac{\mathrm{TP}}{\mathrm{TP}+\mathrm{FN}}\\&\mathrm{Precision~(P)}:P=\frac{\mathrm{TP}}{\mathrm{TP}+\mathrm{FP}}\\&\mathrm{F1~Score~(F1)}:F=\frac{2\times P\times R}{P+R}
\end{align}

\section{The prompt templates used in this paper}
\label{prompt_template}

\onecolumn

\begin{tcolorboxfloat}[!t]
    \begin{tcolorbox}[colback=white, 
                      colframe=darkgray, 
                      width=\textwidth, 
                      arc=2mm, auto outer arc, 
                      breakable, 
                      title={Data Example (Chinese)}]
        \begin{CJK}{UTF8}{gkai} 
            \textcolor[RGB]{180, 101, 5}{【病例ID】}: 63682 
            
            \textcolor[RGB]{180, 101, 5}{【科室】}: 耳鼻咽喉科 
            
            \textcolor[RGB]{180, 101, 5}{【病历摘要】}: 

            \textcolor{blue}{基本信息}: 女，55岁。
            
            \textcolor{blue}{主诉}: 咽痛伴呼吸费力、吞咽困难1天。
            
            \textcolor{blue}{现病史}: 患者1天前无明显诱因出现咽部疼痛，无咳嗽、咳痰，无头痛、发热，无胸闷、心慌等其他不适。患者未予治疗，未行特殊处理。昨日夜间患者无明显诱因出现呼吸费力，吞咽感困难，咽部疼痛加重，自行口服消炎药物未见明显缓解，无意识障碍，无咳嗽，咳痰，无头痛、发热，无恶心、呕吐等其他不适。患者今为求进一步治疗，特来我科就诊，门诊拟急性会厌炎收治入院。入院患者自起病来精神软，饮食睡眠差，大小便未见明显异常。
            
            \textcolor{blue}{既往史}: 自诉糖尿病病史三年，甲状腺肿大，子宫肌瘤病史，日常规律服用阿卡波糖片三次每日，一粒每次。自诉餐前空腹血糖7，餐后血糖小于10，具体不详。早前曾行卵巢囊肿摘除术，自诉现一般情况可。自诉规律服用左旋甲状腺素片半片每天，既往常有早饭过晚感胃痛，日常有反酸病史，未正规就诊及规律用药，无肝炎结核等传染病史，无高血压，冠心病等病史，无外伤无输血史，无食物药物过敏史，按计划预防接种。
            
            \textcolor{blue}{辅助检查}: 

            \underline{血常规示}: 中性细胞比率0.751↑、淋巴细胞比率0.185↓、嗜酸性粒细胞$0.04×10^9/L$、血小板压积0.31; 
            \underline{血生化示}: LDL胆固醇$3.81mol/L$↑、C反应蛋白$28.6mg/L$、总胆固醇$5.84mol/L$、葡萄糖$6.32mmol/L$、糖化血清白蛋白0.1681; 
            \underline{尿常规示}: 白细胞$25×10^9/L$↑、尿潜血$25/ \mu L$; 
            \underline{甲功5项示}: 促甲状腺素$0.184\mu IU/ml$↓; 
            \underline{咽拭子脓液培养示}: 生长正常菌群; 
            \underline{胸片示}: 两肺未见明显活动性病变; 
            \underline{副鼻窦CT示}: 副鼻窦CT平扫未见明显异常，口咽部右侧粘膜稍增厚，请结合临床; 
            \underline{腹部B超示}: 胆囊底部絮状回声，考虑泥沙样结石，肝脏、胰腺、脾脏未见明显占位性病变; 
            \underline{甲状腺彩超示}: 甲状腺多发结节。其他未见明显异常。
            
            \textcolor[RGB]{180, 101, 5}{【临床诊断】}

            \textcolor{blue}{初步诊断}: 急性会厌炎，咽部水肿，2型糖尿病

            \textcolor{blue}{诊断依据}: 患者症状为咽痛伴呼吸费力、吞咽困难1天，查体咽部悬雍垂及软腭水肿，双侧扁桃体1度，咽后壁淋巴滤泡充血，间接喉镜下观：会厌下榻，舌面黏膜水肿呈球状，充血明显，双侧声带窥不清。患者糖尿病病史3年。

            \textcolor{blue}{鉴别诊断}: 根据患者症状.体征，检查等可初步诊断为急性会厌炎，咽部水肿，但亦不排除以下可能：1.会厌囊肿：常可见会厌舌面囊肿样物质隆起，表面光滑成球状，患者常异物感明显，无咽痛，无呼吸，吞咽等困难。2.会厌脓肿：查体会厌表面可见粘膜明显隆起，肿物内可见脓性分泌物流出，患者常感咽喉部疼痛，咽喉部异物感明显。  
            
            \textcolor{blue}{诊断结果}: 急性会厌炎，咽部水肿，右侧扁桃体周围脓肿，2型糖尿病
            
        \end{CJK}
    \end{tcolorbox}
    \captionof{figure}{A data example from CMEMR.} 
    \label{fig:example_zh} 
\end{tcolorboxfloat}

\onecolumn

\begin{tcolorboxfloat}[!t]
    \begin{tcolorbox}[colback=white, 
                      colframe=darkgray, 
                      width=\textwidth, 
                      arc=2mm, auto outer arc, 
                      breakable, 
                      title={Data Example (English)}]

        \textcolor[RGB]{180, 101, 5}{[Case ID]}: 63682 
        
        \textcolor[RGB]{180, 101, 5}{[Department]}: Otorhinolaryngology 
        
        \textcolor[RGB]{180, 101, 5}{[Case Summary]}:
        
        \textcolor{blue}{Basic Information}: Female, 55 years old.
        
        \textcolor{blue}{Chief Complaint}: Sore throat accompanied by labored breathing and difficulty swallowing for 1 day.
        
        \textcolor{blue}{History of Present Illness}: 
        One day ago, the patient developed a sore throat without obvious triggers, accompanied by no cough, sputum, headache, fever, chest tightness, or palpitations. She did not seek treatment or take special measures. Last night, labored breathing and difficulty swallowing emerged, with worsening throat pain. Self-administered anti-inflammatory medication was ineffective. She denied loss of consciousness, nausea, vomiting, or other symptoms. Due to worsening symptoms, she visited our department and was admitted with a preliminary diagnosis of acute epiglottitis. since the onset of illness, the patient has experienced lethargy, poor appetite, and sleep disturbances, with no significant abnormalities in urination or defecation.

        \textcolor{blue}{Past Medical History}:
        The patient reports a 3-year history of diabetes, thyroid enlargement, and uterine fibroids. She takes acarbose regularly (1 tablet, three times daily) and levothyroxine sodium (half a tablet daily). Fasting blood glucose is self-reported at 7 mmol/L, and postprandial glucose under 10 mmol/L, specifics unclear. She underwent ovarian cystectomy and reports stable health. She occasionally experiences gastric pain after delayed breakfast and has a history of acid reflux but has not sought formal treatment. She denies infectious diseases, hypertension, coronary artery disease, trauma, or blood transfusions. No known allergies; vaccinations are up-to-date.

        \textcolor{blue}{Auxiliary Examinations}: 
        
        \underline{Complete Blood Count}: Neutrophil ratio 0.751↑, lymphocyte ratio 0.185↓, eosinophils $0.04×10^9/L$, plateletcrit 0.31; 
        \underline{Biochemical Panel}: LDL cholesterol $3.81 mol/L$↑, C-reactive protein $28.6 mg/L$, total cholesterol $5.84 mol/L$, glucose $6.32 mmol/L$, glycated albumin 0.1681; 
        \underline{Urinalysis}: Leukocytes $25×10^9/L$↑, urine occult blood $25/\mu L$; 
        \underline{Thyroid Function Test (5 items)}: Thyroid-stimulating hormone $0.184 \mu IU/ml$↓; 
        \underline{Throat Swab Pus Culture}: Normal flora growth; 
        \underline{Chest X-ray}: No significant active pulmonary lesions; 
        \underline{Paranasal Sinus CT}: No significant abnormalities in the sinuses; slight mucosal thickening on the right side of the oropharynx; clinical correlation suggested; 
        \underline{Abdominal Ultrasound}: Flocculent echo at the gallbladder fundus, suggestive of sludge-like stones; liver, pancreas, and spleen unremarkable; 
        \underline{Thyroid Ultrasound}: Multiple nodules in the thyroid gland. No other significant abnormalities.
        
        \textcolor[RGB]{180, 101, 5}{[Clinical Diagnosis]}

        \textcolor{blue}{Preliminary Diagnosis}: Acute epiglottitis, pharyngeal edema, Type 2 Diabetes Mellitus (T2DM)
        
        \textcolor{blue}{Diagnostic Basis}:
        The patient presents with a sore throat, labored breathing, and difficulty swallowing for 1 day. Examination reveals uvular and soft palate edema, Grade 1 bilateral tonsils, pharyngeal lymphoid follicular hyperemia, and indirect laryngoscopy showing epiglottic collapse with globular mucosal swelling and marked hyperemia. The patient has a 3-year history of diabetes.
        
        \textcolor{blue}{Differential Diagnosis}:
        Based on symptoms, signs, and examinations, the preliminary diagnosis is acute epiglottitis with pharyngeal edema. Differential considerations include:
        1. **Epiglottic Cyst**: Typically characterized by cystic elevation on the lingual surface of the epiglottis, smooth and globular appearance, often with prominent foreign body sensation but without throat pain, respiratory or swallowing difficulties.
        2. **Epiglottic Abscess**: Physical examination reveals significant mucosal elevation with purulent discharge from the lesion, accompanied by pronounced throat pain and foreign body sensation.
        

        \textcolor{blue}{Final Diagnosis}: 
        Acute epiglottitis, pharyngeal edema, right peritonsillar abscess, Type 2 Diabetes Mellitus.

    \end{tcolorbox}
    \captionof{figure}{A data example from CMEMR(translated).} 
    \label{fig:example_en} 
\end{tcolorboxfloat}

\twocolumn

\begin{CJK}{UTF8}{gkai}
\begin{table*}[!h]
\centering

\begin{tabular}{p{0.9\textwidth}}

\toprule
\hline
\textcolor{blue}{\textbf{Error Class 1: Overlapping Disease Characteristics}}\\
\hline
\textbf{[Case 1]:} Overlapping characteristics of different diseases\\
\textbf{[Key Info]:} Fine scales visible on back rash; prominent petechiae on both lower limbs; fluocinolone acetonide cream\\
\textbf{[Label]:} \underline{Chronic Lichenified Pityriasis}\\
\textbf{[Candidate Diseases]:} Psoriasis, Eczema, Pityriasis Rosea, Henoch-Schönlein Purpura\\
\textbf{[LLM Final Decision]:} Psoriasis (\textcolor{darkred}{\ding{56}}), Pityriasis Rosea (\textcolor{darkred}{\ding{56}})\\

\hline
\textbf{[Case 2]:} Overlapping characteristics within the same category\\
\textbf{[Key Info]:} Fatigue; melena; fecal occult blood test positive (colloidal gold method)\\
\textbf{[Label]:} Gastric cancer with bleeding, \underline{Iron-deficiency anemia}, ...\\
\textbf{[Candidate Diseases]:}
Gastric cancer, Gastrointestinal bleeding, Anemia, ..., \textcolor{gray}{\sout{Iron-deficiency anemia}} (reprioritized, ranking >$topn$)...\\
\textbf{[LLM Final Decision]:} Gastric cancer (\textcolor{darkgreen}{$\checkmark$}), Gastrointestinal bleeding (\textcolor{darkgreen}{$\checkmark$}), Anemia (\textcolor{darkgreen}{$\checkmark$})\\

\hline
\midrule

\textcolor{blue}{\textbf{Error Class 2: Misunderstanding of Examination Indicators}}\\
\hline
\textbf{[Case 1]:} Insufficient understanding of the meaning and utility of examination indicators\\
\textbf{[Key Info]:} EF 27\%; FS 12\%; LAM light chain M-protein positive\\
\textbf{[Label]:} Multiple Myeloma, \underline{Cardiac Amyloidosis}\\
\textbf{[Candidate Diseases]:} Multiple Myeloma, Chronic Heart Failure, ...\\
\textbf{[LLM Final Decision]:} Multiple Myeloma (\textcolor{darkgreen}{$\checkmark$}), Chronic Heart Failure (\textcolor{darkred}{\ding{56}})\\

\hline
\midrule

\textcolor{blue}{\textbf{Error Class 3: Gap Between LLM Reasoning and Physician Diagnostic Labeling}}\\

\hline
\textbf{[Case 1]:} Diagnostic results include previous medical history\\
\textbf{[Key Info]:} Deep ulcer in the middle of the gastric body; white ulcer scars in the duodenal bulb; pulmonary tuberculosis for over a year, already cured\\
\textbf{[Label]:} Gastric ulcer, Duodenal bulb ulcer, ..., \underline{Inactive Pulmonary Tuberculosis}\\
\textbf{[Candidate Diseases]:} Gastric ulcer, Duodenal bulb ulcer, ..., \textcolor{gray}{\sout{Inactive Pulmonary Tuberculosis}} (not considered as a candidate disease)\\
\textbf{[LLM Final Decision]:} Gastric ulcer (\textcolor{darkgreen}{$\checkmark$}), Duodenal bulb ulcer (\textcolor{darkgreen}{$\checkmark$})\\
\hline
\textbf{[Case 2]:} Diagnostic results include abbreviations\\
\textbf{[Key Info]:} Paroxysmal abdominal pain; urine glucose 3+, ketone 2+; normal serum amylase\\
\textbf{[Label]:} \underline{FCPD}, \underline{DKA}\\
\textbf{[Candidate Diseases]:} Diabetes, Chronic Pancreatitis, \underline{DKA}\\
\textbf{[LLM Final Decision]:} Diabetes (\textcolor{darkgreen}{$\checkmark$}), Chronic Pancreatitis (\textcolor{darkred}{\ding{56}})\\

\hline
\bottomrule

\end{tabular}
\caption{Examples of Classification Errors. For clarity, only part of the key information of the selected samples is presented. "\textbf{[Candidate Diseases]}" denotes the set of candidate diseases obtained after the Path-based Reranking stage. Diseases with underlines indicate that diagnoses annotated by the doctor being either misdiagnosed or missed by LLMs. "\textcolor{darkgreen}{$\checkmark$}" indicates that the final decision of LLMs is consistent or partially consistent with the doctor's annotation, while "\textcolor{darkred}{\ding{56}}" indicates the opposite.}

\label{error_analysis}
\end{table*}
\end{CJK}

\begin{figure*}[t]
\centering
\includegraphics[scale=0.45]{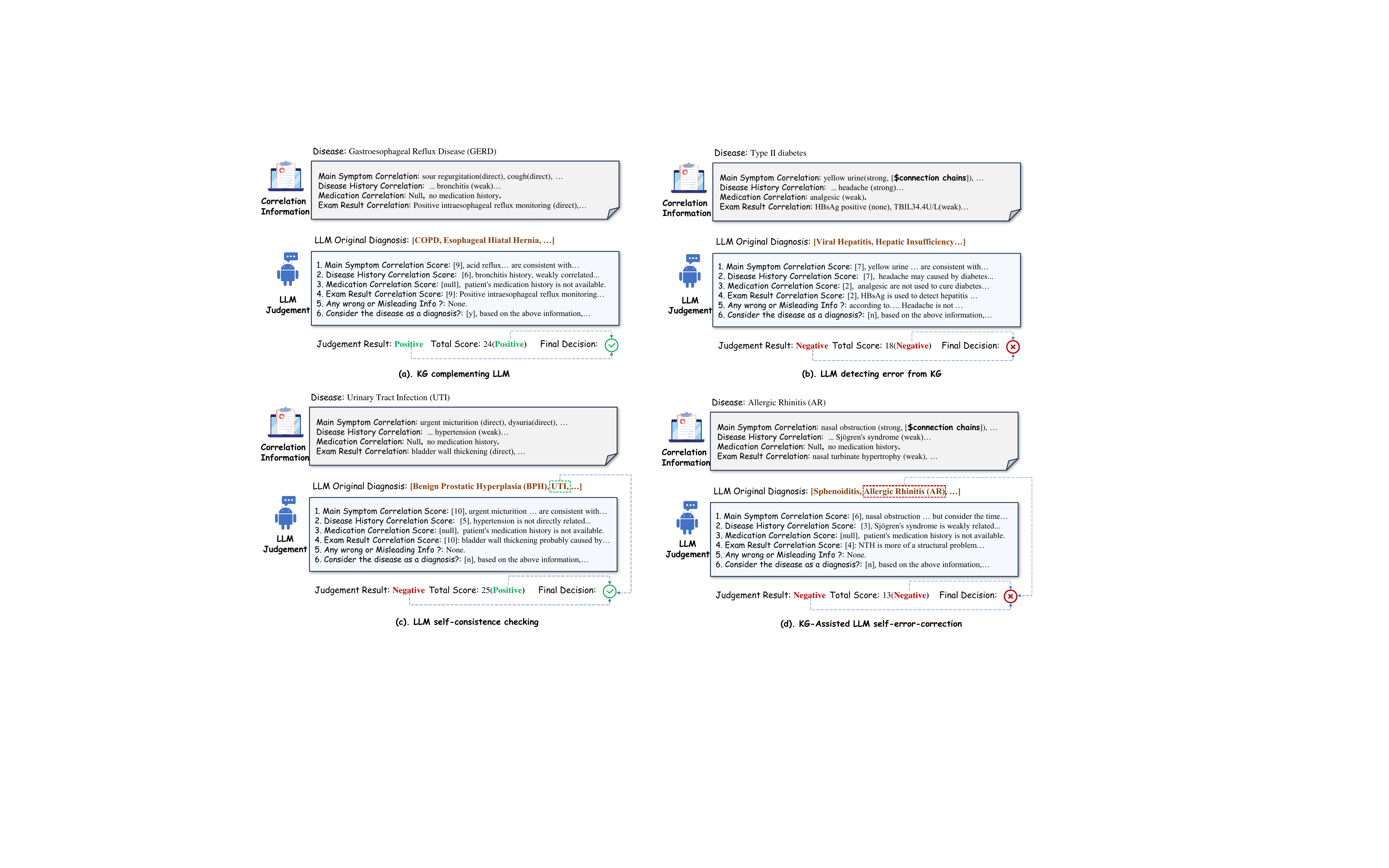}
\caption{Case study.}
\label{case_study}
\end{figure*}

\begin{table*}[b]
\centering
\begin{tabular}{p{0.9\textwidth}}
\toprule
\hline
[\textbf{Role}]<\textcolor{blue}{SYS}>\\
You are an outstanding AI medical expert. You can summarize critical information for diagnosis based on the content of the patient's medical records.\\
\hline
[\textbf{Role}]<\textcolor{blue}{USR}>\\
Below is a portion of the electronic medical record of a real patient. Please read the following content carefully to understand the patient's basic condition. \\

\#\# Patient Medical Record Content \\

$\text{"""}$\\
"History of Present Illness": \$\{HPI\}\\
"Past Medical History": \$\{PMH\}\\
$\text{"""}$\\

\#\# Task: \\
Based on the above content, please summarize the key information useful for diagnosis and treatment and generate a summary report. \\

\#\# Report Format Requirements: \\
Please fill in the "[]" sections according to the following format to complete the report. Use concise language whenever possible. \\
$\text{"""}$\\
1. Main symptoms: []\\
2. Recent medical visits: [] (if none, write "none")\\
3. Past medical history: [] (if none, write "none")\\
4. Past surgical history: [] (if none, write "none")\\
5. Medication usage: [] (if none, write "none")\\
$\text{"""}$\\

\#\# Output: \\
\$\{\} \\
\hline
\bottomrule
\end{tabular}
\caption{The default prompt for the LLM Summarization module~(for the patients' basic condition) .}
\label{summary_prompt}
\end{table*}

\begin{table*}[h]
\centering
\begin{tabular}{p{0.9\textwidth}}
\toprule
\hline
[\textbf{Role}]<\textcolor{blue}{SYS}>\\
You are an excellent AI medical expert. You can summarize key information useful for diagnosis based on the patient's examination results.\\
\hline
[\textbf{Role}]<\textcolor{blue}{USR}>\\
\#\# Task:\\
Please summarize and generalize the key information useful for diagnosis based on the patient's examination results.\\
\#\# Example\\
$\text{"""}$\\
$\text{[Patient's Examination Results]}$\\
"Physical Examination": Bilateral waistline symmetry, no tenderness in the bilateral ureteral regions, bladder area distended, no palpable mass, no redness or abnormal discharge at the urethral opening, no abnormalities in the scrotum, and no abnormalities in the bilateral testicles and epididymis. Digital rectal exam: Prostate approximately 4.0×5.0cm in size, soft, central area slightly shallow, small nodules palpable.\\
"Laboratory and Aided Examination": Ultrasound results show 1. Bilateral kidney cysts 2. Prostatic hyperplasia 3. No abnormalities in the ureters and bladder.\\
$\text{"""}$\\
Please refer to the above example to summarize the patient's examination results.\\
$\text{[Patient's Examination Results]}$ \\
"Physical Examination": \$\{PE\}\\
"Laboratory and Aided Examination": \$\{LAE\}\\
\#\#Output: \\
\$\{\}\\
\hline
\bottomrule
\end{tabular}
\caption{The default prompt for the LLM Summarization module~(for the patients' exam results).}
\label{summary_exam_prompt}
\end{table*}

\begin{table*}[h]
\centering
\begin{tabular}{p{0.9\textwidth}}
\toprule
\hline
[\textbf{Role}]<\textcolor{blue}{SYS}>\\
You are an outstanding AI medical expert. You can perform a preliminary disease diagnosis based on the patient's condition.\\
\hline
[\textbf{Role}]<\textcolor{blue}{USR}>\\
\#\#Patient Information\\
$\text{"""}$\\
\text{[General Condition]}: \$\{summary\_1\}\\
\text{[Examination Findings]}: \$\{summary\_2\}\\
$\text{"""}$\\

\#\#Task\\
Based on the patient's symptoms, medical visit history, past medical history, and examination results, predict the possible diseases the patient may have (you can provide the top-\$\{n\} possible predictions). Please only output the prediction results, do not output any other content.

\#\#Prediction Results\\
Predicted Disease 1: \$\{\} \hspace*{2em} Predicted Disease 2: \$\{\}\hspace*{2em}  Predicted Disease 3: \$\{\} ...\\
\hline
\bottomrule
\end{tabular}
\caption{The default prompt for the LLM Direct Diagnose Module.}
\label{diagnosis_prompt}
\end{table*}

\begin{table*}[h]
\centering
\begin{tabular}{p{0.9\textwidth}}
\toprule
\hline
[\textbf{Role}]<\textcolor{blue}{SYS}>\\
You are an experienced medical expert. You can evaluate the reasonableness of existing diagnostic results by considering the patient's symptoms, medical history, medication usage, and examination results.\\
\hline
[\textbf{Role}]<\textcolor{blue}{USR}>\\
\#\#Patient Information\\
$\text{"""}$\\
\text{[General Condition]}: \$\{summary\_1\}\\
\text{[Examination Findings]}: \$\{summary\_2\}\\
$\text{"""}$\\
A doctor has made a preliminary diagnosis based on the above information, with the diagnosis being: \$\{disease\}\\
You need to consider whether this diagnosis is correct. To do this, you queried a medical knowledge graph and obtained the following information:\\
\#\#Correlation Information\\
$\text{"""}$\\
Correlation between diagnosis \$\{disease\} and patient's main symptoms:
\$\{correlation\_1\}\\
Correlation between diagnosis \$\{disease\} and patient's medical history: 
\$\{correlation\_2\}\\
Correlation between diagnosis \$\{disease\} and patient's medication usage: 
\$\{correlation\_3\}\\
Correlation between diagnosis \$\{disease\} and patient's examination results: 
\$\{correlation\_4\}\\
$\text{"""}$\\

\#\#Task\\
Based on the patient's condition and the above information, and in combination with your own knowledge, please quantitatively evaluate the reasonableness of the diagnosis \$\{disease\}.\\

\#\#Requirements\\
$\text{"""}$\\
1.Consistency with the patient's chief complaint score: [?]~(out of 10)\\
2.Correlation with the patient's medical history score: [?]~(out of 10)\\
3.Correlation with the patient's medication usage score: [?]~(out of 10)\\
4.Correlation with the patient's examination results score: [?]~(out of 10)\\
5.Are there any errors or misleading information in the "Correlation Information" section~?\\
6.Can this disease be used as a diagnostic result: [?]~(y/n)\\
$\text{"""}$\\
\#\#Output:\\
\$\{\}\\
\hline
\bottomrule
\end{tabular}
\caption{The default prompt for the LLM Diagnosis Evaluation Module.}
\label{analysis}
\end{table*}

\end{document}